\documentclass[conference]{IEEEtran}
\IEEEoverridecommandlockouts

\usepackage{cite}
\usepackage{amsmath,amssymb,amsfonts}
\usepackage{algorithmic}
\usepackage{graphicx}
\usepackage{textcomp}
\usepackage{xcolor}
\usepackage{multirow}
\usepackage{booktabs}
\usepackage{url}
\usepackage{siunitx}
\usepackage{xcolor}
\usepackage{colortbl}
\def\BibTeX{{\rm B\kern-.05em{\sc i\kern-.025em b}\kern-.08em
    T\kern-.1667em\lower.7ex\hbox{E}\kern-.125emX}}

\definecolor{lightgray}{RGB}{242,243,245}
\definecolor{lightblue}{RGB}{237,245,255}
\definecolor{lightgreen}{RGB}{227,250,227}
\definecolor{lightorange}{RGB}{255,243,217}
\definecolor{lightpurple}{RGB}{251,240,255}

\begin{document}

\title{Geolocation-Aware Robust Spoken Language Identification}

\author{\IEEEauthorblockN{Qingzheng Wang, Hye-jin Shim, Jiancheng Sun, and Shinji Watanabe} \IEEEauthorblockA{Carnegie Mellon University\\
qingzhew@andrew.cmu.edu, shimhz6.6@gmail.com, jianches@andrew.cmu.edu, shinjiw@ieee.org}
}

\maketitle

\begin{abstract}
While Self-supervised Learning (SSL) has significantly improved Spoken Language Identification (LID), existing models often struggle to consistently classify dialects and accents of the same language as a unified class. To address this challenge, we propose geolocation-aware LID, a novel approach that incorporates language-level geolocation information into the SSL-based LID model. Specifically, we introduce geolocation prediction as an auxiliary task and inject the predicted vectors into intermediate representations as conditioning signals. This explicit conditioning encourages the model to learn more unified representations for dialectal and accented variations. Experiments across six multilingual datasets demonstrate that our approach improves robustness to intra-language variations and unseen domains, achieving new state-of-the-art accuracy on FLEURS (97.7\%) and 9.7\% relative improvement on ML-SUPERB~2.0 dialect set.
\end{abstract}


\section{Introduction}

Spoken language identification (LID) is becoming increasingly essential as speech technology expands toward multilingual scalability.
With the emergence of speech foundation models trained on hundreds or even thousands of languages~\cite{pratap2024mms, chen2024xeus, babu2021xls, radford2023whisper, peng2024owsm, zhang2023google}, accurately identifying the language of an utterance has become a critical first step in both dataset curation pipelines and runtime systems.
For instance, LID enables language-aware automatic speech recognition (ASR) by routing input to the appropriate language-specific module~\cite{pratap2024mms}, and supports large-scale multilingual dataset construction through filtering and annotation~\cite{valk2021voxlingua107, barrault2023seamless, barrault2023seamlessm4t, anonymous2025owsmctcv4}.

Recent advances in self-supervised learning (SSL) have improved the robustness and cross-lingual transferability of speech representations, which can be fine-tuned for LID with high accuracy~\cite{pratap2024mms, chen2024xeus, liu2022efficient}.
Prior studies have shown that SSL models predominantly capture phonetic representations~\cite{choi24b_interspeech, yang23v_interspeech}, making them particularly effective for distinguishing languages with distinct sound patterns. 

However, dialects and accents within the same language often differ significantly in phonetic representations, which can lead to misclassifications of these intra-language variations as another language. 
For instance, English encompasses a wide range of regional dialects and accents, such as American and Indian English, which differ phonetically despite sharing the same language identity.
One potential solution is to assign fine-grained dialect or accent labels for classification, but it is incompatible with most downstream tasks such as ASR and speech translation, which operate at the language level and expect to generalize across dialectal and accented variations.

\begin{table}[t]
    \centering
    \caption{Accuracy (\%) with joint prediction of language ID and meta features from \texttt{lang2vec}~\cite{littell2017uriel}. \colorbox{lightorange}{Orange}/\textbf{Bold}: best overall.}
    \label{tab:preliminary}
    \begin{tabular}{lcc}
    \toprule
        \multirow{2.5}{*}{Meta Info} & \multicolumn{2}{c}{ML-SUPERB~2.0} \\
 \cmidrule(lr){2-3}
 & Dev & Dialect \\
    \midrule
        LID-only & 89.0 & 73.4 \\
        \rowcolor{lightorange}
        Geolocation & \textbf{89.5} & \textbf{73.8} \\
        Inventory & 88.8 & 68.2 \\
        Phonology & 88.8 & 73.6 \\
        Syntax & 88.9 & 67.2 \\
    \bottomrule
    \end{tabular}
    \vspace*{-4mm}
\end{table}

To address this challenge, we explore using language-level meta information as auxiliary supervision to guide the model to learn unified representations for dialectal and accented variations.
Among several candidates, including geolocation, phonology, phonetic inventory, and syntax, we compare their effectiveness by predicting each as an auxiliary task jointly with LID.
Our preliminary results (Table~\ref{tab:preliminary}) with the ML-SUPERB 2.0~\cite{ml-superb2_interspeech2025} show that geolocation provides the most consistent improvement, suggesting that it can serve as a strong signal to unify intra-language variations.

Motivated by this finding, we propose geolocation-aware LID, a novel framework that incorporates language-level geolocation information into SSL-based LID models. 
Specifically, we introduce geolocation prediction as an auxiliary task at both intermediate layers of the SSL encoder and the downstream embedding extractor. 
Predicted geolocation vectors from intermediate layers are injected into subsequent layers as conditioning signals, encouraging the model to develop more compact and consistent representations for dialectal and accented speech within the same language.

Our key contributions are as follows: (i) we propose geolocation-aware LID, a new approach that incorporates geolocation prediction and conditioning into the SSL-based LID model; (ii) we empirically demonstrate the effectiveness of language-level geolocation signals in improving robustness to intra-language variations; (iii) we develop a robust LID system supporting 157 languages, achieving \textbf{new state-of-the-art (SOTA)} accuracy with relative improvements of 0.5\% on FLEURS (97.7\%)~\cite{conneau2023fleurs}, and 2.0\% and 9.7\% on ML-SUPERB 2.0~\cite{ml-superb2_interspeech2025} development (88.6\%) and dialect (86.8\%) set, respectively.
Relevant code, model weights (including our SOTA checkpoint), and training logs are publicly available.\footnote{\url{https://github.com/espnet/espnet/tree/master/egs2/geolid/lid1}}

\section{Related Studies}

\subsection{Geographic Information for LID and Speech Processing}

The integration of geographic information into spoken language identification remains unexplored. 
Foley et al.~\cite{foley2024you} explored \textit{utterance-level} speech geolocation prediction as a proxy task to improve LID, showing that geolocation-pretrained encoders yield better performance than directly fine-tuned SSL models.
To our knowledge, this is the only work on using geolocation information for spoken language identification.
In the field of textual language identification, Dunn et al.~\cite{dunn-edwards-brown-2024-geographically} showed similar benefits by incorporating geographic priors into region-specific LID models.
More broadly, geographic information has been leveraged in ASR via geolocation vectors for dialect modeling~\cite{cao2021improving} and location-aware language models for local vocabulary~\cite{xiao2018geographic}.
In this work, we extend this line of research by predicting \textit{language-level} geolocation and injecting the predicted geolocation as conditioning information into SSL representations to improve spoken LID performance. 

\subsection{Intermediate Layer Prediction and Conditioning}

Prediction at intermediate layers has proven effective for regularizing training in ASR models. For example, applying Connectionist Temporal Classification (CTC) loss to encoder layers~\cite{lee2021intermediate, dejavu} and adding LID-aware CTC loss in SSL encoder layers~\cite{wang2025lidctc} have been used.
While auxiliary prediction tasks provide useful training signals, conditioning intermediate representations on these predictions allows subsequent layers to explicitly use these signals.
For instance, self-conditioned CTC~\cite{nozaki21_interspeech} conditioned final predictions on intermediate layer predictions to relax the conditional independence assumption.
Chen et al.~\cite{chen2023lidctc} extended this by conditioning intermediate layers on LID predictions to improve multilingual ASR performance.
Beyond the ASR scope, Lu et al.~\cite{lusslr} leveraged language- and speaker-specific information extracted from intermediate layers to adapt the pretrained SSL encoder. 
However, conditioning on geolocation information has not yet been explored.
In this paper, we propose to condition SSL encoder intermediate layers on geolocation predictions.

\section{Geolocation-Aware LID}

To enhance robustness against dialectal and accented variations, we extend the SSL-based LID framework by incorporating geolocation information.
As shown in Fig.~\ref{fig:model}, our architecture builds on the conventional SSL-based LID pipeline, consisting of a pretrained upstream SSL encoder, a downstream language embedding extractor, and a classification head.
We introduce an auxiliary geolocation prediction task at both intermediate layers of the SSL encoder and the output of the embedding extractor. 
To enable the SSL encoder to directly utilize geolocation information, we inject intermediate-layer geolocation predictions as conditioning signals into subsequent encoder layers. 

\textbf{\begin{figure}[t]
  \centering
  \includegraphics[width=0.7\linewidth]{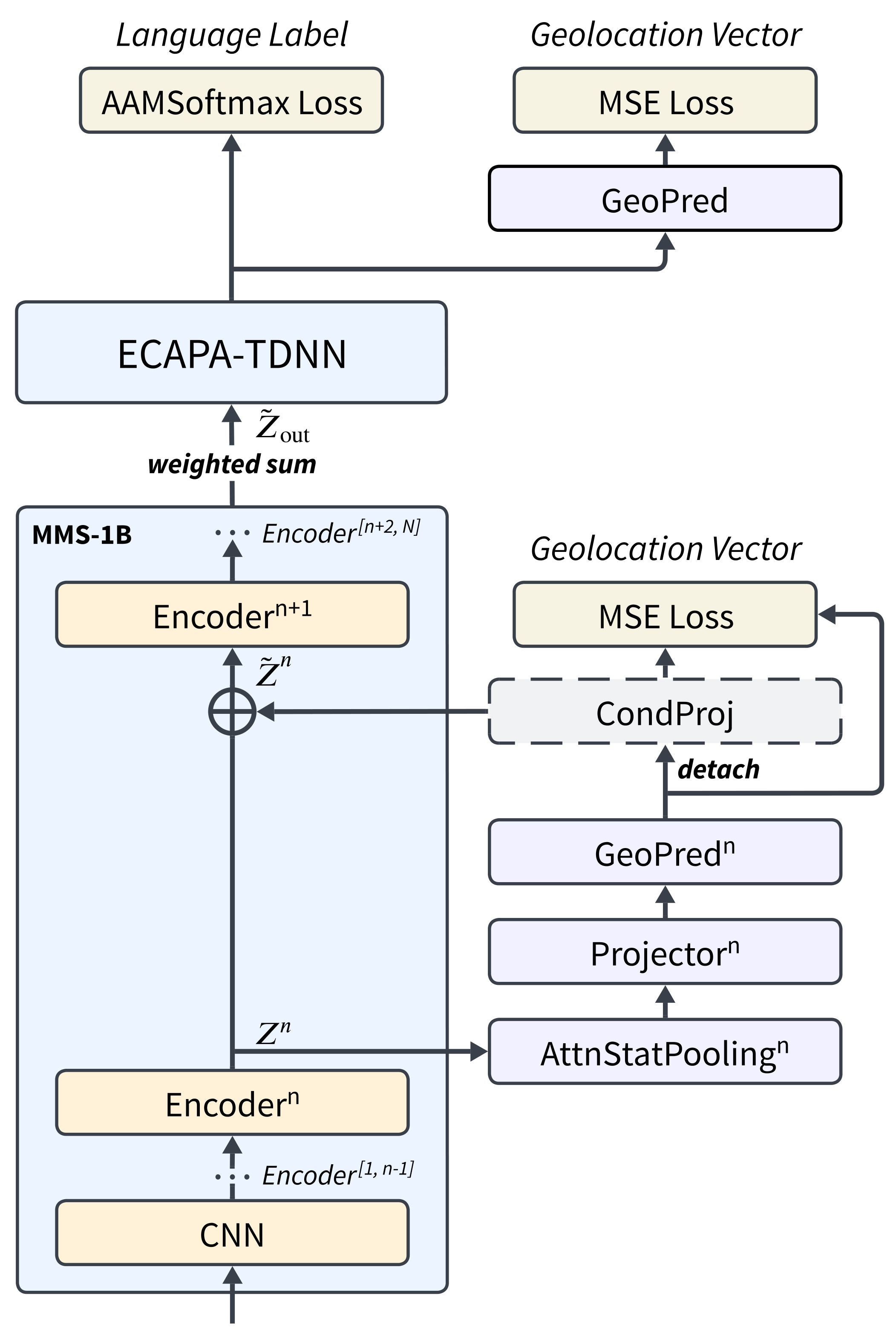}
  \caption{Overview of the proposed geolocation-aware LID architecture. Geolocation vectors are predicted from a set of selected intermediate layers and the downstream embedding extractor. Intermediate predictions are detached and re-injected into the encoder via a conditioning projection module (dashed block), with design choices (shared vs. independent, frozen vs. trainable) depending on layer positions. A weighted sum of all hidden states of encoder layers is passed to ECAPA-TDNN for embedding extraction.}
  \label{fig:model}
  \vspace*{-4mm}
\end{figure}}

\subsection{SSL-based LID Framework}
\label{sec:baseline}

In this section, we describe the core architecture of our SSL-based LID model (left side of Fig.~\ref{fig:model}). 
We use MMS-1B~\cite{pratap2024mms} as the SSL encoder, a 1B-parameter model based on wav2vec 2.0~\cite{baevski2020wav2vec} pretrained on over 1,400 languages (see large blue block in Fig.~\ref{fig:model}). 
The sub-branches extending from MMS-1B for geolocation conditioning will be described in Section~\ref{sec:geocond}.

Given a raw audio input, the model first applies a convolutional waveform encoder (the CNN block in Fig.~\ref{fig:model}) that extracts a $T$-length sequence of $D$-dimensional acoustic features $X \in \mathbb{R}^{T \times D}$.
These features are then processed by a stack of $N$-layer Transformer encoders~\cite{vaswani2017transformer} $\{\text{Encoder}^n\}^N_{n=1}$ (yellow blocks in Fig.~\ref{fig:model}; layers
$\text{Encoder}^{[1,n-1]}$ and $\text{Encoder}^{[n+2,N]}$ are omitted for clarity):
\begin{align}
    \label{ssl_encoder}
    Z^n = \text{Encoder}^n(Z^{n-1}),
\end{align}
where $Z^n = (\mathbf{z}_{t}^n \in \mathbb{R}^{D} | t = 1, \dots, T)$ is the $n$-th layer output, with $Z^0 = X$.
The final SSL encoder output $Z_{\text{out}}$ is obtained through a \textit{weighted sum} of all encoder hidden states~\cite{matthew2018featurizer, yang21s3prl}:
\begin{align}
    \label{ssl_output}
    Z_{\text{out}} = \sum_{n=0}^{N}\alpha^n Z^{n},
\end{align}
where $\alpha$ are learnable parameters satisfying $\sum_{n=0}^{N}\alpha^n = 1$.

The aggregated SSL representation $Z_{\text{out}}$ is then processed by ECAPA-TDNN~\cite{desplanques20_interspeech}, followed by MMS-1B in Fig.~\ref{fig:model}, to extract language embeddings. 
This module processes frame-level features through a series of ECAPA blocks, which incorporate 1-D convolutional layers and squeeze-and-excitation Res2Blocks~\cite{senet, gao2019res2net}:
\begin{align}
\label{ecapa}
H = \text{ECAPA-Blocks}(Z_{\text{out}}),
\end{align}
where $H \in \mathbb{R}^{T' \times C}$ with $C$ channels and $T'$ frames after convolutions.
Then, the frame-level features are aggregated using attentive statistics pooling~\cite{desplanques20_interspeech, okabe18_interspeech}:
\begin{align}
    \mathbf{s} = \text{AttnStatPooling}(H),
\end{align}
where $\mathbf{s} \in \mathbb{R}^{2C}$ contains the pooled mean and standard deviation statistics.
Finally, the pooled statistics are projected to obtain the language embedding $\mathbf{e} \in \mathbb{R}^{E}$:
\begin{align}
    \label{project_embedding}
    \mathbf{e} = \text{Projector}(\mathbf{s}),
\end{align}
where the projector includes batch normalization~\cite{ioffe2015batch} followed by a linear transformation.

We adopt the AAMSoftmax~\cite{deng2019arcface} loss function enhanced with the sub-center technique~\cite{zhao2021speakin}, as implemented in ESPnet-SPK~\cite{jung2024espnet}, to perform language classification (see the top of Fig.~\ref{fig:model}):
\begin{align}
    \label{loss_class}
    \mathcal{L}_{\text{class}} = \text{AAMSoftmax}(\mathbf{e}, y; K, m, s),
\end{align}
where $y$ is the ground-truth language label, $K$ is the number of sub-centers capturing intra-class variations, $m$ is the angular margin, and $s$ is the scaling factor.

\subsection{Geolocation Vectors}
\label{sec:geolocation_vector}

To utilize geographic information, we use the geolocation vectors provided by the \texttt{lang2vec} project~\cite{littell2017uriel} to represent the abstract geolocation of each language.
These vectors are derived from estimated geographic coordinates of languages, obtained from typological resources like Glottolog~\cite{glottolog2025}.
The coordinates of each language are transformed into vectors by computing normalized great-circle distances to 299 uniformly distributed reference points on Earth (generated via a spherical Fibonacci lattice~\cite{gonzalez2010measurement}).
The resulting 299-dimensional vectors with values between $[0, 1]$ provide a continuous and structured encoding that is well-suited for both prediction tasks and integration into high-dimensional hidden spaces.

\subsection{Geolocation Prediction as an Auxiliary Task}
\label{sec:geopred}

To guide the model to learn language-discriminative representations, we incorporate an auxiliary geolocation prediction task into the fine-tuning process. 
Given a speech utterance in language $l$ with ground-truth geolocation vector $\mathbf{v}_l$, we predict the geolocation vector from the language embedding $\mathbf{e}$ in \eqref{project_embedding}:
\begin{align}
    \label{down_geo_vector}
    \hat{\mathbf{v}}_l = \text{GeoPred}(\mathbf{e}),
\end{align}
where $\text{GeoPred}(\cdot)$ is a linear projection module (upper-right block in Fig.~\ref{fig:model}). 
The geolocation prediction loss is defined as:
\begin{align}
    \label{loss_geo}
    \mathcal{L}_{\text{geo}} = \text{MSE}(\hat{\mathbf{v}}_l, \mathbf{v}_l),
\end{align}
where $\text{MSE}$ denotes the mean squared error loss. 
We combine the classification loss in \eqref{loss_class} and $\mathcal{L}_{\text{geo}}$ as:
\begin{align}
    \label{loss1}
    \mathcal{L}_1 = (1 - \lambda)\mathcal{L}_{\text{class}} + \lambda\mathcal{L}_{\text{geo}},
\end{align}
where $\lambda \in [0,1]$ balances the classification and geolocation prediction objectives.

\subsection{Conditioning the SSL Encoder on Geolocation Predictions}
\label{sec:geocond}

While geolocation prediction provides explicit supervision for LID, its output is not directly incorporated into the SSL representation.
To enable the SSL encoder to explicitly use the geolocation information, we inject geolocation conditioning signals into the intermediate layers of the SSL encoder. 

We select a subset of intermediate layers $\mathcal{M} \subseteq \{1, \dots, N\}$ from the SSL encoder defined in \eqref{ssl_encoder}.
For each selected layer $n \in \mathcal{M}$, the frame-level hidden states $Z^n$, introduced in \eqref{ssl_encoder}, are processed to obtain intermediate language embeddings and geolocation predictions:
\begin{align}
    \label{inter_pool_proj}
    \mathbf{e}^n &= \text{Projector}^n(\text{AttnStatPooling}^n(Z^n)), \\
    \label{inter_pred}
    \hat{\mathbf{v}}_{l}^{n} &= \text{GeoPred}^n(\mathbf{e}^n),
\end{align}
where all modules are layer-specific and correspond to the purple blocks in the right sub-branches of Fig.~\ref{fig:model}.
Unlike $\mathbf{e}$ in \eqref{project_embedding} and $\hat{\mathbf{v}}_l$ in \eqref{down_geo_vector} extracted from the downstream module, $\mathbf{e}^n$ and $\hat{\mathbf{v}}_{l}^{n}$ capture distinct characteristics at each depth.

As each dimension of the geolocation vector encodes the distance to a fixed reference point, the geolocation vector is numerically sensitive: slight perturbations in its values can shift the implied geolocation. 
To prevent distortion by gradients from the downstream classification objective in \eqref{loss_class}, we detach the predicted geolocation vector into $\bar{\mathbf{v}}_{l}^{n}$ before projecting it into the conditioning signal $\mathbf{c}^n$:
\begin{align}
    \label{detach}
    \bar{\mathbf{v}}_{l}^{n} &= \text{detach}(\hat{\mathbf{v}}_{l}^{n}), \\
    \label{condition}
    \mathbf{c}^n &= \text{CondProj}(\bar{\mathbf{v}}_{l}^{n}),
\end{align}
where $\text{CondProj}$ is a linear layer (the dashed block in Fig.~\ref{fig:model}) and $\mathbf{c}^n \in \mathbb{R}^D$.
This detachment only blocks the gradient from the $\text{CondProj}$ layer; the original $\hat{\mathbf{v}}_{l}^{n}$ remains connected to the computational graph and is supervised by the intermediate-layer geolocation loss for layer $n$:
\begin{align}
    \label{loss_inter_geo}
    \mathcal{L}_{\text{geo}}^n = \text{MSE}(\hat{\mathbf{v}}_{l}^{n}, \mathbf{v}_l).
\end{align}
Therefore, the geolocation prediction modules in \eqref{inter_pool_proj} and \eqref{inter_pred} are optimized only by the intermediate-layer geolocation objective.
The effect of detachment will be shown in Section~\ref{results:ablation_downloss_detach}. 

As the sole interface between the geolocation predictions and the SSL encoder, the design of $\text{CondProj}$ in \eqref{condition} plays a crucial role in shaping how geolocation signals are represented and utilized.
This module can be configured to be either shared or independent across layers, and either frozen or trainable during fine-tuning.
Shared vs. independent controls whether the geolocation signal is tailored for each layer, while frozen vs. trainable determines whether it remains fixed or is adaptively modulated. 
As no configuration is universally optimal, we empirically evaluate these design choices in Section~\ref{results:condproj_design}.

The geolocation conditioning signal is then added to each frame of the hidden states (see the $\oplus$ operation in Fig.~\ref{fig:model}):
\begin{align}
    \tilde{\mathbf{z}}_t^n = \mathbf{z}_t^n + \mathbf{c}^n,
\end{align}
forming the conditioned representation $\tilde{Z}^n = (\tilde{\mathbf{z}}_{t}^n \in \mathbb{R}^{D} | t = 1, \dots, T)$ that serves as input to the subsequent layer. 
With the conditioning signals injected into the selected layers $n \in \mathcal{M}$, the final SSL encoder output in~\eqref{ssl_output} becomes:
\begin{align}
    \tilde{Z}_{\text{out}} = \sum_{n \notin \mathcal{M}}\alpha_n Z^n + \sum_{n \in \mathcal{M}}\alpha_n \tilde{Z}^n,
\end{align}
resulting in geolocation-aware SSL representations.

Given the classification loss $\mathcal{L}_{\text{class}}$~\eqref{loss_class}, downstream geolocation loss $\mathcal{L}_{\text{geo}}$~\eqref{loss_geo}, and intermediate-layer geolocation losses $\mathcal{L}_{\text{geo}}^n$ for layers $n \in \mathcal{M}$~\eqref{loss_inter_geo}, 
the overall loss is defined as:
\begin{align}
    \label{loss2}
    \mathcal{L}_2 = \left(1 - \lambda\right)\mathcal{L}_{\text{class}} + \lambda \left(\left(1 - \gamma\right)\mathcal{L}_{\text{geo}} + \gamma \frac{\sum_{n\in\mathcal{M}}\mathcal{L}_{\text{geo}}^n}{|\mathcal{M}|}\right),
\end{align}
where $\gamma \in [0, 1]$ balances the downstream and intermediate-layer geolocation prediction losses.

\section{Experiments}

\subsection{Datasets}

We primarily train our models on VoxLingua107~\cite{valk2021voxlingua107} with 6,628-hour 107-language YouTube recordings and evaluate both on the development set of VoxLingua107 and five out-of-domain datasets to show generalization capability: Babel~\cite{gales2014speech}, FLEURS~\cite{conneau2023fleurs}, VoxPopuli~\cite{wang-etal-2021-voxpopuli}, and the development and dialect development sets of ML-SUPERB~2.0~\cite{ml-superb2_interspeech2025}. 
Table~\ref{tab:datasets} summarizes all datasets used in our experiments.
For each, we evaluate only on languages that overlap with the VoxLingua107 training set.\footnote{Babel: development utterances longer than 10s; FLEURS: official test split; VoxPopuli: development set of transcribed speech; ML-SUPERB~2.0: follows setup in the ML-SUPERB~2.0 challenge~\cite{mlsuperb2025_website}.}
Therefore, the number of evaluated languages is often smaller than the official test set size listed in Table~\ref{tab:datasets}.
We further train our models on the combined training sets of all five datasets (9,865 hours, 157 languages) to improve domain coverage and upper-bound performance.\footnote{Babel: utterances longer than 10s from the full-language-pack training set; ML-SUPERB~2.0: same processing as evaluation; VoxPopuli: transcribed training set; others use official splits.}

\subsection{Model Configuration}

We use the 1B-parameter MMS model\footnote{\url{https://huggingface.co/facebook/mms-1b}} as the upstream SSL encoder, which consists of 48 Transformer layers with hidden size $D=1280$ (see \eqref{ssl_encoder}). 
The encoder is fully fine-tuned during training. 
The downstream ECAPA-TDNN uses channel size $C=512$ (see \eqref{ecapa}), and the language embedding dimension is $E=192$ (for both downstream \eqref{project_embedding} and intermediate \eqref{inter_pool_proj}). 
The AAMSoftmax loss in \eqref{loss_class} is applied with $K=3$ sub-centers, margin $m=0.5$, and scaling factor $s=30$.

To determine the optimal layers for geolocation conditioning, we experiment with four layer selection $\mathcal{M}$ strategies (see Section~\ref{sec:geocond}): bottom $\{0, 4, 8, 12\}$, middle $\{16, 20, 24, 28\}$, top $\{32, 36, 40, 44\}$, and full $\{0, 4, 8, \dots, 44\}$, denoted as 0-12, 16-28, 32-44, and 0-44, respectively. 
In addition, we perform ablation on the conditioning projection module in \eqref{condition}, comparing (i) shared vs. independent projections across layers, and (ii) frozen vs. trainable parameters.

\subsection{Training Setup}

For combined training, we use a tri-stage learning rate schedule~\cite{ott2019fairseq} with warmup 5k steps from \num{6e-6} to \num{1e-5}, hold for 20k, then decay to \num{1e-6} over 75k. 
Gradient accumulation is applied every 2 steps (VoxLingua107-only) or 4 steps (combined), with batch sizes of 3min and 1.5min, respectively. 
Optimization uses Adam~\cite{Kingma2014AdamAM} with $\beta_1=0.9$, $\beta_2=0.98$. 
We apply balanced data sampling~\cite{pratap2024mms} with upsampling factor $\beta_{\text{lang}} = 0.5$ for languages, and $\beta_{\text{dataset}} = 0.3$ for datasets in combined training.
We tune $\lambda$ and $\gamma$ in loss $\mathcal{L}_1$ \eqref{loss1} and $\mathcal{L}_2$ \eqref{loss2} over predefined sets with 0.2 and 0.4 selected, respectively. 
Ablation variants include setting $\gamma{=}1$ (see \eqref{loss2}) and removing $\text{detach}(\cdot)$ (see \eqref{detach}).
For inference, we use the highest-accuracy checkpoint on the VoxLingua107 development set for VoxLingua107-only training, and the 62k-step checkpoint for the combined training.
All experiments use ESPnet~\cite{watanabe2018espnet} with S3PRL~\cite{yang21s3prl} and run on one NVIDIA~H200.

\begin{table}[t]
\centering
\setlength{\tabcolsep}{0.3pt}
\renewcommand{\arraystretch}{0.98}
\caption{Overview of datasets used in experiments. VL107-only: train on VoxLingua107 only; Combined: train on all training sets; (137, 8): dev and dialect-dev sets in ML-SUPERB~2.0; Seen/Unseen: whether the dataset is used during fine-tuning.}
\vspace{-1mm}
\label{tab:datasets}
\begin{tabular}{lcccccc}
\toprule
\multirow{2.5}{*}{Dataset} & \multirow{2.5}{*}{Domain} & \#Langs. & \multirow{2.5}{*}{Dialect} & \multicolumn{2}{c}{Training Setup} \\
\cmidrule(lr){3-3} \cmidrule(lr){5-6}
& & Train/Test & & VL107-only & Combined \\
\midrule
VoxLingua107~\cite{valk2021voxlingua107} & YouTube & 107/33 & No & Seen & Seen \\
Babel~\cite{gales2014speech} & Telephone & 25/25 & No & Unseen & Seen \\
FLEURS~\cite{conneau2023fleurs} & Read speech & 102/102 & No & Unseen & Seen \\
ML-SUPERB 2.0~\cite{ml-superb2_interspeech2025} & Mixed & 137/(137, 8) & Yes & Unseen & Seen \\
VoxPopuli~\cite{wang-etal-2021-voxpopuli} & Parliament & 16/16 & No & Unseen & Seen \\
\bottomrule
\end{tabular}
\vspace*{-4mm}
\end{table}

\begin{table*}[t]
\centering
\renewcommand{\arraystretch}{0.98}
\caption{Accuracy (\%) of models trained on VoxLingua107 across in-domain and out-of-domain test sets. Geo Pred: downstream geolocation prediction only; Geo Cond: intermediate-layer geolocation conditioning with downstream geolocation prediction; Macro Avg.: macro average accuracy over all sets; Indep.: Independent; Train.: Trainable; \underline{Underlined}: group best; \textbf{Bold}: best per column; \colorbox{lightgray}{Gray}: baseline; \colorbox{lightpurple}{Purple}: macro avg. outperforms baseline; \colorbox{lightorange}{Orange}: best overall.}
\vspace{-1mm}
\begin{tabular}{llccccccccc}
\toprule
\multirow{3.5}{*}{\#} & \multirow{3.5}{*}{Model} & \multirow{3.5}{*}{Layers} & \multirow{3.5}{*}{CondProj Type} & \multicolumn{1}{c}{In-domain} & \multicolumn{5}{c}{Out-of-domain} & \multirow{3.5}{*}{Macro Avg.} \\ \cmidrule(lr){5-5} \cmidrule(lr){6-10}
 & & & & \multirow{2.5}{*}{VoxLingua107} & \multirow{2.5}{*}{Babel} & \multirow{2.5}{*}{FLEURS} & \multicolumn{2}{c}{ML-SUPERB 2.0} & \multirow{2.5}{*}{VoxPopuli} & \\
 \cmidrule(lr){8-9}
  & & & & & & & Dev & Dialect & & \\
\midrule
\rowcolor{lightgray}
1 & Baseline & – & None & \underline{94.2} & \underline{86.7} & \underline{\textbf{95.8}} & 89.0 & 73.4 & 85.6 & 87.5 \\
2 & Geo Pred & – & None & 94.1 & 86.0 & 95.6 & \underline{89.5} & \underline{73.8} & \underline{88.9} & \cellcolor{lightpurple}\underline{88.0} \\
\midrule
\multicolumn{11}{c}{\textit{Conditioning Projection Design and Position}} \\
3 & Geo Cond & 0-12 & Indep. + Frozen & 94.3 & \underline{85.9} & \underline{95.1} & 89.1 & \underline{73.9} & \underline{90.6} & \cellcolor{lightpurple}\underline{88.1} \\
4 & Geo Cond & 0-12 & Indep. + Train. & \underline{94.5} & 85.1 & 93.1 & 88.9 & 73.5 & 87.5 & 87.1 \\
5 & Geo Cond & 0-12 & Shared + Frozen & 94.0 & 83.6 & 94.7 & \underline{89.4} & 72.4 & 90.4 & 87.4 \\
6 & Geo Cond & 0-12 & Shared + Train. & 94.4 & 85.2 & 93.5 & 88.0 & 71.7 & 88.4 & 86.9 \\
\rowcolor{white}
\multicolumn{11}{c}{} \\[-2ex]

7 & Geo Cond & 16-28 & Indep. + Frozen & \underline{95.0} & 85.9 & 93.2 & 89.0 & \underline{76.3} & \underline{89.0} & \cellcolor{lightpurple}\underline{88.1} \\
8 & Geo Cond & 16-28 & Indep. + Train. & 94.3 & 84.7 & 92.1 & 88.2 & 72.5 & 85.9 & 86.3 \\
9 & Geo Cond & 16-28 & Shared + Frozen & 94.0 & \underline{86.2} & \underline{94.7} & 88.7 & 74.6 & 87.1 & 87.5 \\
10 & Geo Cond & 16-28 & Shared + Train. & 94.5 & 86.1 & 94.5 & \underline{89.4} & 71.3 & 88.3 & 87.3 \\
\rowcolor{white}
\multicolumn{11}{c}{} \\[-1.5ex]

11 & Geo Cond & 32-44 & Indep. + Frozen & 94.2 & 87.1 & \underline{95.0} & 89.0 & 77.2 & \underline{90.4} & \cellcolor{lightpurple}88.8 \\
12 & Geo Cond & 32-44 & Indep. + Train. & 93.7 & 85.3 & 93.7 & 88.3 & 70.3 & 86.5 & 86.3 \\
13 & Geo Cond & 32-44 & Shared + Frozen & 94.3 & 85.9 & 94.3 & 88.8 & \underline{\textbf{80.7}} & 89.2 & \cellcolor{lightpurple}88.8 \\
\rowcolor{lightorange}
14 & Geo Cond & 32-44 & Shared + Train. & \underline{94.9} & \underline{\textbf{87.7}} & 93.5 & \underline{89.3} & 78.8 & 89.5 & \cellcolor{lightpurple}\underline{\textbf{88.9}} \\
\rowcolor{white}
\multicolumn{11}{c}{} \\[-1.5ex]
15 & Geo Cond & 0-44 & Indep. + Frozen & 93.9 & 83.5 & 94.9 & \underline{\textbf{89.7}} & \underline{76.5} & \underline{\textbf{91.2}} & \cellcolor{lightpurple}\underline{88.3} \\
16 & Geo Cond & 0-44 & Indep. + Train. & 93.7 & 84.8 & 94.0 & 88.3 & 72.7 & 87.9 & 86.9 \\
17 & Geo Cond & 0-44 & Shared + Frozen & \underline{94.4} & 83.9 & \underline{95.0} & 89.1 & 68.8 & 89.9 & 86.8 \\
18 & Geo Cond & 0-44 & Shared + Train. & 93.9 & \underline{86.5} & 94.3 & 88.2 & 70.4 & 87.3 & 86.8 \\
\midrule
\multicolumn{11}{c}{\textit{Remove Downstream Geolocation Loss ($\gamma=1$)}} \\
19 & Geo Cond & 32-44 & Shared + Train. & \textbf{95.2} & 86.8 & 93.2 & 88.5 & 77.4 & 90.2 & \cellcolor{lightpurple}88.6 \\
\multicolumn{11}{c}{\textit{Remove }$\text{detach}(\cdot)$} \\
20 & Geo Cond & 32-44 & Shared + Train. & 94.2 & 85.2 & 93.8 & 89.4 & 73.8 & 87.2 & 87.2 \\
\bottomrule
\end{tabular}
\label{tab:crossdomain_results}
\vspace*{-4mm}
\end{table*}

\section{Results on VoxLingua107-only Training}
\label{sec:reseult_vl107}

Table~\ref{tab:crossdomain_results} presents the LID accuracy of models trained on VoxLingua107 and evaluated on both in-domain and out-of-domain test sets.
Three settings are compared: (i) a baseline model without geolocation supervision (Section~\ref{sec:baseline}), (ii) a model with downstream geolocation prediction (Section~\ref{sec:geopred}), and (iii) models with geolocation conditioning on intermediate layers (Section~\ref{sec:geocond}).
Overall, both geolocation prediction and conditioning models outperform the baseline (see purple-highlighted macro averages in Table~\ref{tab:crossdomain_results}).
The geolocation conditioning model with shared, trainable projections on layers 32-44 achieves the highest macro accuracy of 88.9\%, outperforming both the baseline and geolocation prediction-only models. 
This demonstrates the effectiveness of injecting geolocation conditioning signals into intermediate representations.
The most significant improvements occur on challenging sets such as ML-SUPERB~2.0 dialect and VoxPopuli, with absolute improvements of 7.3\% and 5.6\% respectively, suggesting that geolocation conditioning signals improve robustness to both intra-language variations and domain shifts.
Performance on FLEURS slightly declines, but remains comparable to the baseline, introducing minimal trade-off.

\subsection{Design and Position of Conditioning Projection}
\label{results:condproj_design}

\textbf{Early-layer conditioning benefits from independent and frozen projection modules.} 
Conditioning early layers (0–12, 16–28) performs best with independent and frozen projection modules.
At layers 0–12, the independent frozen projection achieves up to 3.1\% higher accuracy than its trainable counterpart on VoxPopuli.
This result implies that frozen projection modules provide more consistent conditioning and stabilize low-level features than trainable modules.
Among frozen settings, independent projections outperform shared ones (e.g., 88.1\% vs. 87.5\%)
, highlighting the benefits of layer-specific integration of geolocation cues.

\textbf{Deep-layer representations offer a stable semantic space for geolocation conditioning.}
Deep-layer conditioning (layers 32–44) benefits more from shared and trainable projections, which achieves the highest macro average accuracy (88.9\%) among all configurations.
Notably, shared and frozen projections remain competitive, especially on the ML-SUPERB~2.0 dialect development set, scoring the best accuracy of 80.7\%.
These results indicate that deep layers provide semantically stable representations suitable for both static and adaptive conditioning.
Furthermore, shared projections consistently outperform independent ones on macro average accuracy, implying that a unified transformation better supports geolocation integration at deep layers.

\textbf{Conditioning across all layers does not yield cumulative performance gains.}
Applying geolocation conditioning across all layers (0-44) mirrors early-layer trends: frozen projections work better than trainable ones, with the independent frozen setup achieving the highest accuracy in ML-SUPERB~2.0 development set (89.7\%).
However, this approach underperforms compared to deep-layer injection (32-44), and in some cases (e.g., independent trainable), even falls short of early-layer injection (e.g., macro average accuracy 86.9\% vs. 87.1\% at layers 0-12).
This implies that broad conditioning may introduce redundancy rather than cumulative benefit.

\begin{table*}[t]
\centering
\caption{Accuracy (\%) of representative LID models. Type: SSL-based, acoustic feature-based, joint LID-ASR, geolocation-pretrained, and our geolocation-conditioned LID model (layers 32–44, shared trainable projection). Macro Avg.: average over all sets. XEUS: ML-SUPERB~2.0 results from~\cite{mlsuperb2025_website}. Ours: VoxLingua107-only (VL107-only) or combined training. \textbf{Bold}: best overall.}
\vspace{-1mm}
\renewcommand{\arraystretch}{0.98}
\label{tab:overall}
\begin{tabular}{llccccccc}
\toprule
\multirow{2.5}{*}{Model} & \multirow{2.5}{*}{Type} & \multirow{2.5}{*}{VoxLingua107} & \multirow{2.5}{*}{Babel} & \multirow{2.5}{*}{FLEURS} & \multicolumn{2}{c}{ML-SUPERB 2.0} & \multirow{2.5}{*}{VoxPopuli} & \multirow{2.5}{*}{Macro Avg.}\\ 
 \cmidrule(lr){6-7}
 & & & & & Dev & Dialect & & \\
\midrule
MMS-LID-4017~\cite{pratap2024mms} & SSL  & 93.9 & – & 97.2  & – & – & – & – \\
XLS-R-attentive~\cite{kukk22_interspeech} & SSL & \textbf{95.3} & – & – & – & – & – & – \\
TitaNet-LID~\cite{jia23b_interspeech} & Acoustic & 94.4 & – & – & – & – & – & – \\
XEUS~\cite{chen2024xeus} & LID-ASR & – & – & 93.0 & 77.1 & 79.1 & – & – \\
MMS 1B LIDCTC~\cite{wang2025lidctc} & LID-ASR & – & – & –  & 86.9 & 74.2 & – & – \\
OWSM v4 medium~\cite{anonymous2025owsmctcv4} & LID-ASR & – & – & 95.6 & – & – & – & – \\
Geo 1B~\cite{foley2024you} & Geo Pretrain & – & – & 96.7 & – & – & – & – \\
\midrule
Ours (VL107-only) & Geo Cond & 94.9 & 87.7 & 93.5 & \textbf{89.3} & 78.8 & 89.5 & 88.9 \\
Ours (Combined) & Geo Cond & 94.4 & \textbf{95.4} & \textbf{97.7} & 88.6 & \textbf{86.8} & \textbf{99.0} & \textbf{93.7} \\
\bottomrule
\end{tabular}
\vspace*{-4mm}
\end{table*}

\subsection{Effect of Downstream Geolocation Loss and Detachment}
\label{results:ablation_downloss_detach}

To assess the effect of downstream geolocation loss, we remove it by setting $\gamma = 1$ in \eqref{loss2}, while keeping intermediate-layer geolocation conditioning (experiment 19 in Table~\ref{tab:crossdomain_results}).
Compared to experiment 14, the performance drops in out-of-domain settings, despite achieving the best in-domain accuracy on VoxLingua107 (95.2\%).
This suggests that downstream geolocation supervision benefits cross-domain generalization.

We further examine the role of detaching the intermediate geolocation prediction before projecting it into the hidden space. 
Removing the $\text{detach}(\cdot)$ operation leads to a significant performance degradation (see experiment 20), especially on the ML-SUPERB~2.0 dialect development set (5.0\% absolute drop compared to experiment 14).
This indicates that without detachment, gradients from the classification objective interfere with the learning of geolocation vectors, causing them to align with the classification target rather than preserving the geolocation information.

\begin{table}[t]
\centering
\small
\setlength{\tabcolsep}{2.3pt}
\renewcommand{\arraystretch}{0.98}
\caption{Accuracy (\%) on ML-SUPERB 2.0 dialect dev set. Geo Cond: layers 32–44 with shared frozen projection; \underline{Underlined}: best across both settings.}
\vspace{-1mm}
\label{tab:dialect}
\begin{tabular}{lccccccccc|c}
\toprule
Model & ara & deu & ell & eng & guj & spa & tam & tel \\
\midrule
Baseline & \underline{65.3} & 76.5 & 75.5 & 67.4 & \underline{99.0} & 96.4 & \underline{100.0} & \underline{98.0} \\
Geo Cond & 61.5 & \underline{88.3} & \underline{83.4} & \underline{76.5} & 97.9 & \underline{98.5} & \underline{100.0} & \underline{98.0} \\
\bottomrule
\end{tabular}
\vspace*{-4mm}
\end{table}

\subsection{Improvement on Dialectal and Accented Variations}
\label{results:dialect_accent}

Table~\ref{tab:dialect} presents detailed results for each language in the ML-SUPERB~2.0 dialect development set.
Geolocation conditioning significantly improves or preserves accuracy on most languages with dialectal or accented variations, except for Arabic (ara).
This suggests that geolocation conditioning improves the model's robustness to intra-language variations consistently across languages.

\begin{figure}[t]
    \centering
    \includegraphics[width=1\linewidth]{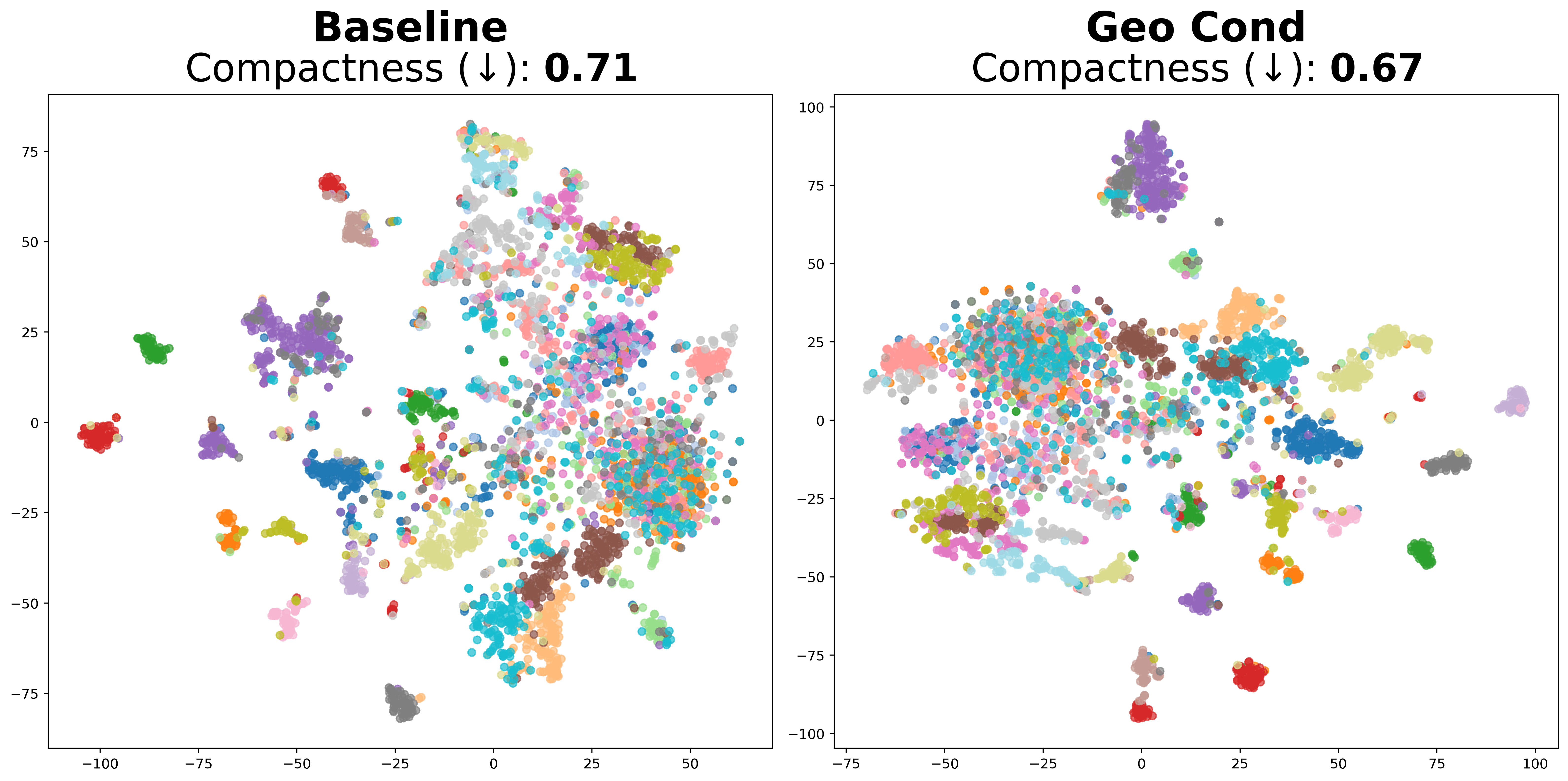}
    \caption{t-SNE plots of English speech embeddings from ML-SUPERB~2.0 dialect dev set. Colors indicate accents within the English class. Geo Cond: geolocation-conditioned model (layers 32–44, shared frozen). Compactness: average distance to the English embedding centroid, lower indicates tighter clustering.}
    \label{fig:tsne}
    \vspace*{-4mm}
\end{figure}

To further analyze its effect on intra-language variations, we visualize the utterance-level embeddings for English speech in ML-SUPERB~2.0 dialect development set in Fig.~\ref{fig:tsne}.
With geolocation conditioning, the compactness score decreases from 0.71 to 0.67, indicating tighter clustering of intra-language embeddings.
This demonstrates that geolocation signals, serving as a unifying constraint, guide the model to learn compact representations for intra-language variations, leading to better generalization across dialects and accents.

\section{Results on Combined Training}
\label{results:overall}

Building on Section~\ref{sec:reseult_vl107}, we expand training data from 6,628 to 9,865 hours with broader domain coverage, and train the geolocation conditioning model using shared, trainable conditioning projections on layers 32-44, achieving SOTA performance.
Table~\ref{tab:overall} reports the LID accuracy of our geolocation-aware LID models compared to existing SOTA systems. 
Our model achieves new SOTA accuracy on FLEURS (97.7\%) and ML-SUPERB~2.0 (dev: 88.6\%, dialect dev: 86.8\%), while maintaining comparable results on VoxLingua107.
Compared with Geo 1B, which relies on utterance-level geolocation pretraining, our method uses only language-level geolocation signals and achieves higher accuracy on FLEURS (97.7\% vs. 96.7\%).
This demonstrates that estimated, language-level geolocation is sufficient to improve LID performance without requiring fine-grained utterance-level location labels.
The checkpoint of our SOTA model is publicly available.

\section{Conclusion}

In this paper, we propose geolocation-aware LID, a novel approach that incorporates language-level geolocation supervision and conditioning into SSL-based LID models.
Using geolocation vectors from \texttt{lang2vec} project~\cite{littell2017uriel}, we predict the language geolocation at both SSL encoder intermediate layers and the downstream embedding extractor, and inject the intermediate-layer predictions as conditioning signals into the encoder.
Experiments show that our approach improves overall model performance, particularly enhancing robustness to dialectal and accented variations.
Trained on a 157-language multi-domain dataset, our model achieves new SOTA results on FLEURS~\cite{conneau2023fleurs} and ML-SUPERB~2.0~\cite{ml-superb2_interspeech2025}.

\section*{Acknowledgment}

Experiments used PSC Bridges2 and NCSA Delta via ACCESS CIS210014 and IRI120008P, supported by NSF grants \#2138259, \#2138286, \#2138307, \#2137603, \#2138296.

\bibliographystyle{IEEEtran}
\bibliography{refs}

\end{document}